\documentclass[10pt,twocolumn,letterpaper]{article}
 
\usepackage[pagenumbers]{cvpr} 

\usepackage{graphicx}
\usepackage{amsmath}
\usepackage{amssymb}
\usepackage{booktabs}
\usepackage{multirow}
\usepackage{arydshln}
\usepackage{color}
\usepackage{bm}
\usepackage{colortbl} 
\usepackage{color}
\usepackage{bm}
\usepackage{bbding}
\usepackage{pifont}
\usepackage{wasysym}

\usepackage{colortbl}  
\usepackage{xcolor}
\usepackage{array} 
\usepackage{wrapfig}  
\usepackage{ulem}
\definecolor{citecolor}{HTML}{229954}
\definecolor{best}{RGB}{225, 225, 225}
\definecolor{second}{RGB}{190, 205, 243}
\definecolor{aaa}{RGB}{222, 222, 230}
\definecolor{bbb}{RGB}{226, 240, 220}

\usepackage[pagebackref,breaklinks,colorlinks]{hyperref}
 
\definecolor{citecolor}{HTML}{229954}
\definecolor{best}{RGB}{225, 225, 225}

\usepackage[capitalize]{cleveref}
\crefname{section}{Sec.}{Secs.}
\Crefname{section}{Section}{Sections}
\Crefname{table}{Table}{Tables}
\crefname{table}{Tab.}{Tabs.}

\def\cvprPaperID{*} 
\def\confName{CVPR}
\def\confYear{2023}

\begin{document}

\title{A Dive into SAM Prior in Image Restoration}

\author{
Zeyu Xiao$^{1*}$  \quad 
Jiawang Bai$^{2*}$ \quad 
Zhihe Lu$^{3}$\thanks{Equal contribution} \quad 
Zhiwei Xiong$^{1}$\\
$^{1}$University of Science and Technology of China\\ 
$^{2}$Tsinghua University \quad
$^{3}$National University of Singapore
\\
}

\maketitle

\begin{abstract} 
The goal of image restoration (IR), a fundamental issue in computer vision, is to restore a high-quality (HQ) image from its degraded low-quality (LQ) observation.
Multiple HQ solutions may correspond to an LQ input in this poorly posed problem, creating an ambiguous solution space.
This motivates the investigation and incorporation of prior knowledge in order to effectively constrain the solution space and enhance the quality of the restored images.
In spite of the pervasive use of hand-crafted and learned priors in IR, limited attention has been paid to the incorporation of knowledge from large-scale foundation models.
In this paper, we for the first time leverage the prior knowledge of the state-of-the-art segment anything model (SAM)~\cite{kirillov2023segment} to boost the performance of existing IR networks in an parameter-efficient tuning manner.
In particular, the choice of SAM is based on its robustness to image degradations, such that HQ semantic masks can be extracted from it.
In order to leverage semantic priors and enhance restoration quality, we propose a lightweight \underline{S}AM \underline{p}rior \underline{t}uning (SPT) unit.
This plug-and-play component allows us to effectively integrate semantic priors into existing IR networks, resulting in significant improvements in restoration quality.
As the only trainable module in our method, the SPT unit has the potential to improve both efficiency and scalability.
We demonstrate the effectiveness of the proposed method in enhancing a variety of methods across multiple tasks, such as image super-resolution and color image denoising.

\end{abstract}
\section{Introduction}

Image restoration (IR) is a fundamental problem in computer vision that aims to recover high-quality (HQ) images from their degraded low-quality (LQ) observations caused by various degradations, such as blur, noise and compression artifacts. 
The IR tasks encompass image super-resolution (SR), image denoising, dehazing, JPEG deblocking, etc.
However, due to the nature of the degradation process, it is an ill-posed problem in practice, leading to multiple HQ solutions corresponding to an LQ input, posing significant challenges for accurate IR.

\begin{figure}[!t]
  \includegraphics[width=\linewidth]{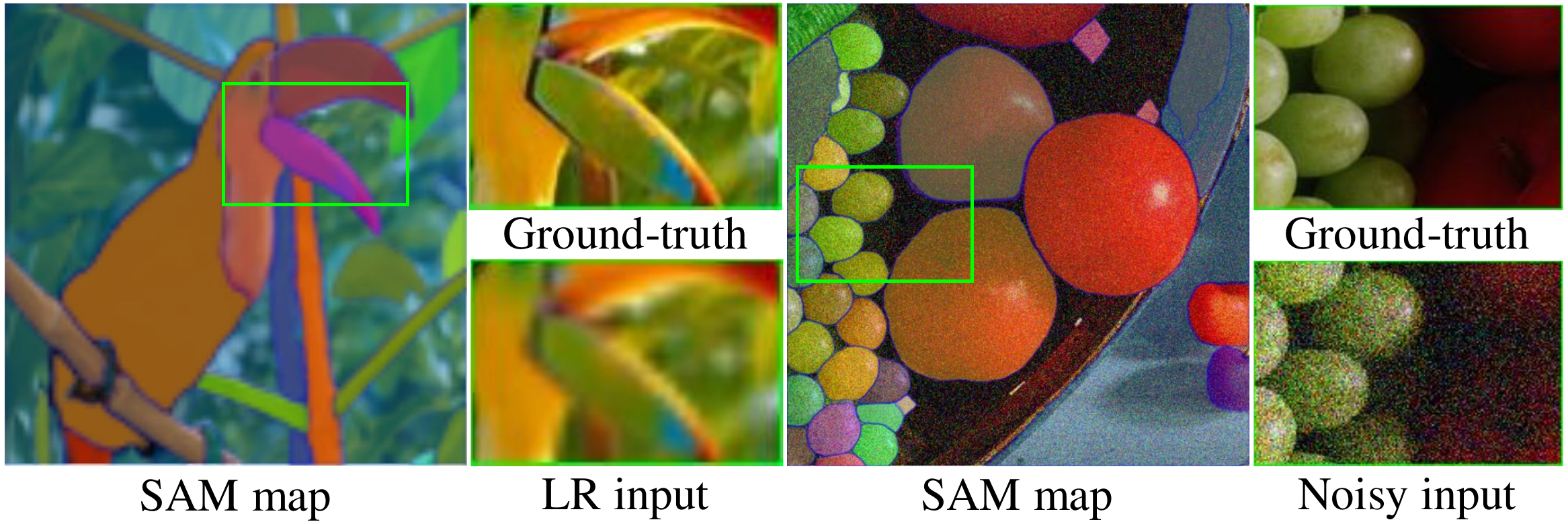}
  \caption{Illustration of SAM's robustness on low-quality images (\textit{e.g.} low-resolution and noisy images).
  It shows SAM can segment objects correctly given low-quality images.
  This observation motivates us to leverage the semantic priors extracted from SAM, a large-scale foundation model, to enhance image restoration performance.
  Examples are from Set5\_Bird and McMaster\_0007, respectively.
  }
  \label{fig:teaser}
\end{figure}

Numerous image priors~\cite{geman1984stochastic,rudin1992nonlinear,zhu1997prior,roth2005fields,he2010single} have been proposed to regularize the solution space of latent clear images in IR tasks.
For instance, the self-similarity prior~\cite{ebrahimi2007solving,glasner2009super,freedman2011image,singh2015super,huang2015single} produces visually pleasing results in image SR task.
Total variation~\cite{rudin1992nonlinear}, wavelet-domain processing~\cite{donoho1995noising}, and BM3D~\cite{dabov2007image} are proposed for the image denoising task by assuming the prior distribution to be smoothness, low rank and self-similarity.
For image dehazing, assumptions are made on atmospheric light, transmission maps, or clear images~\cite{he2010single,fattal2014dehazing}.
While these task-specific image priors have demonstrated superior performance for IR methods, they are frequently based on observations of specific image properties that may not always reflect the inherent image properties.
In addition, the design and selection of task-specific image priors rely on manual and empirical efforts, and the corresponding IR models require intricate optimization.

Recently, it has been increasingly popular to adopt deep models to construct more general priors for IR tasks.
For instance, the seminal work on deep image prior (DIP)~\cite{ulyanov2018deep} has shown that a randomly initialized convolutional neural network (CNN) can implicitly capture texture-level image priors, which can be utilized for IR.
SinGAN\cite{shaham2019singan} demonstrates that a randomly initialized generative adversarial network (GAN) model can capture rich patch statistics after being trained on a single image.
Furthermore, a GAN generator trained on a large dataset of natural images can be used as a generic image prior, referred to as a deep generative prior (DGP)~\cite{pan2020dgp}.
The mentioned methods have shown remarkable performance in IR and image manipulation tasks.
In particular, the CNN and GAN models used in these works are either trained from scratch on a single image or pre-trained on an external dataset.

In this paper, we focus on examining whether foundation models pre-trained on extremely large-scale datasets, such as those containing billions of samples, with strong transfer capabilities can provide richer and more helpful priors for IR tasks.
To this end, we take the first step towards leveraging the semantic-aware prior extracted from a powerful foundation model for segmentation, segment anything model (SAM)~\cite{kirillov2023segment}, which has been trained on a massive dataset called SA-1B containing 1 billion masks and 11 million images.
Our motivation for using SAM as a semantic prior for IR tasks stems from its remarkable robustness on degraded images, including those that are with low-resolution and noise, as illustrated in Figure~\ref{fig:teaser}.
Specifically, we obtain semantic masks of a degraded image by feeding it to the pre-trained SAM, which is referred to as the SAM prior in this paper.
Our method utilizes semantic masks acquired from SAM to enhance the performance of existing IR methods through integration with a lightweight SAM prior tuning (SPT) unit.
This integration of high-level semantic information with intermediate features leads to superior restoration results. 
Specifically, the SPT unit acts as a plug-and-play component by selectively transferring semantic priors to enhance the low-level features and spatial structures of the input LQ image.

To better exploit the potential of the semantic priors obtained from SAM, we propose a parameter-efficient tuning scheme to update the SPT units.
The SPT unit consists of a small number of learnable parameters and can be easily integrated into existing IR methods. 
Our proposed method efficiently integrates semantic priors with existing intermediate features of various CNN-based and Transformer-based IR methods, yielding significant performance improvements over the baselines on benchmark datasets for a range of IR tasks, including image SR and color image denoising.
With the success of the SPT unit in IR tasks, we hope that our work can encourage further studies on incorporating semantic priors into other deep learning-based models.

Overall, our contributions can be summarized as follows:

(1) This paper introduces a novel approach to enhance the performance of IR methods by leveraging the prior knowledge obtained from the state-of-the-art foundation model for segmentation, SAM.
This is the first time such a large-scale pre-trained prior has been used in the context of IR, and we demonstrate that it can be highly effective in improving the restoration quality.

(2) In order to incorporate the semantic priors obtained from SAM, we propose a lightweight SPT unit that can be easily integrated into existing IR methods as a plug-and-play component.
By designing the SPT unit as the only trainable module, we achieve both efficiency and scalability, in contrast to full fine-tuning pipeline which can be computationally expensive and time-consuming.

(3) We comprehensively evaluate the effectiveness of our proposed SPT unit as a plug-in for enhancing existing IR methods, including both CNN-based and Transformer-based methods, on various tasks such as image SR and color image denoising.
Experimental results demonstrate that our method consistently outperforms existing state-of-the-art methods, highlighting its superiority and generalizability.

\section{Related Work}

\subsection{Image Restoration}
Compared to traditional model-based IR methods~\cite{xiong2009image,gu2012fast, timofte2013anchored, timofte2014a, michaeli2013nonparametric, he2010darkchannel}, learning-based methods, particularly those based on CNNs, have shown impressive performance and gained increasing popularity.
These deep models learn mappings between LQ and HQ images from large-scale paired datasets.
Since the pioneering work of SRCNN~\cite{dong2014srcnn} (for image SR), DnCNN~\cite{zhang2017DnCNN} (for image denoising), and ARCNN~\cite{dong2015compression} (for JPEG compression artifact reduction), a large number of CNN-based models have been proposed to improve model representation ability through more elaborate neural network architecture designs, such as residual blocks~\cite{kim2016vdsr, cavigelli2017cas, zhang2021DPIR}, dense blocks~\cite{wang2018esrgan, zhang2018RDN, zhang2020RDNIR}, and others~\cite{chen2016TNRD, lai2017LapSRN, li2018multi,zhang2018srmd, wang2019learning, wang2021unsupervised, wang2021learning, liang2021fkp, liang21hcflow, liang21manet, zhang2018ffdnet, tai2017memnet, isobe2020video, wei2021unsupervised, guo2020closed, cheng2021mfagan, deng2021deep, zhang2019RNAN, peng2019dsnet, jia2019focnet, fu2019jpeg, kim2019pseudo, li2020mdcn,li2020learning,fu2021model,xia2022coarse,li2023learning}.
Some of these models have also incorporated attention mechanisms inside the CNN framework, such as channel attention~\cite{zhang2018rcan, dai2019SAN, niu2020HAN}, non-local attention~\cite{liu2018NLRN, mei2021NLSA}, and adaptive patch aggregation~\cite{zhou2020IGNN}.
Recently, due to the limited ability of CNNs to model long-range dependencies, researchers have started to explore the use of pure self-attention modules for IR~\cite{liang2021swinir,zamir2022restormer,chen2021pre,chen2022cross,zhang2023accurate,lu2021efficient,xia2023diffir}.
In contrast to existing IR methods, our method does not introduce any novel architecture.
Instead, we aim to enhance the performance of existing methods by leveraging the prior generated from a large pre-trained model, such as SAM~\cite{kirillov2023segment}, in a tuning manner, refining and polishing the existing intermediate features through the proposed lightweight SPT unit.

\subsection{Hand-Crafted Image Priors}
Image priors that describe various statistics of natural images have been widely developed and adopted in IR and image editing.
For different IR tasks, priors are also designed specifically based on the characteristics of the imaging and degradation models.
In the image super-resolution (SR) task, the self-similarity prior is able to produce visually pleasing results without extensive training on external databases since a natural image tend to recur within and across scales of the same image~\cite{ebrahimi2007solving,glasner2009super,freedman2011image,singh2015super,huang2015single}.
The heavy-tailed gradient prior~\cite{shan2008high}, sparse kernel prior~\cite{fergus2006removing}, l0 gradient prior~\cite{xu2013unnatural}, normalized sparsity prior~\cite{krishnan2011blind} and dark channel prior~\cite{pan2016blind} are proposed to solve the image deblurring task.
While these traditional hand-crafted priors frequently capture specific statistics and serve specific purposes, there is a growing interest in finding more general priors that capture richer image statistics via deep learning models.
In this paper, we present a parameter-efficient tuning scheme to leverage the prior knowledge from SAM for the task of IR.
To the best of our knowledge, our work is the first to introduce the use of SAM for the task of image restoration, demonstrating the potential of leveraging pre-trained semantic priors for improving IR methods.

\begin{figure*}[t]
\centering
\includegraphics[width=\linewidth]{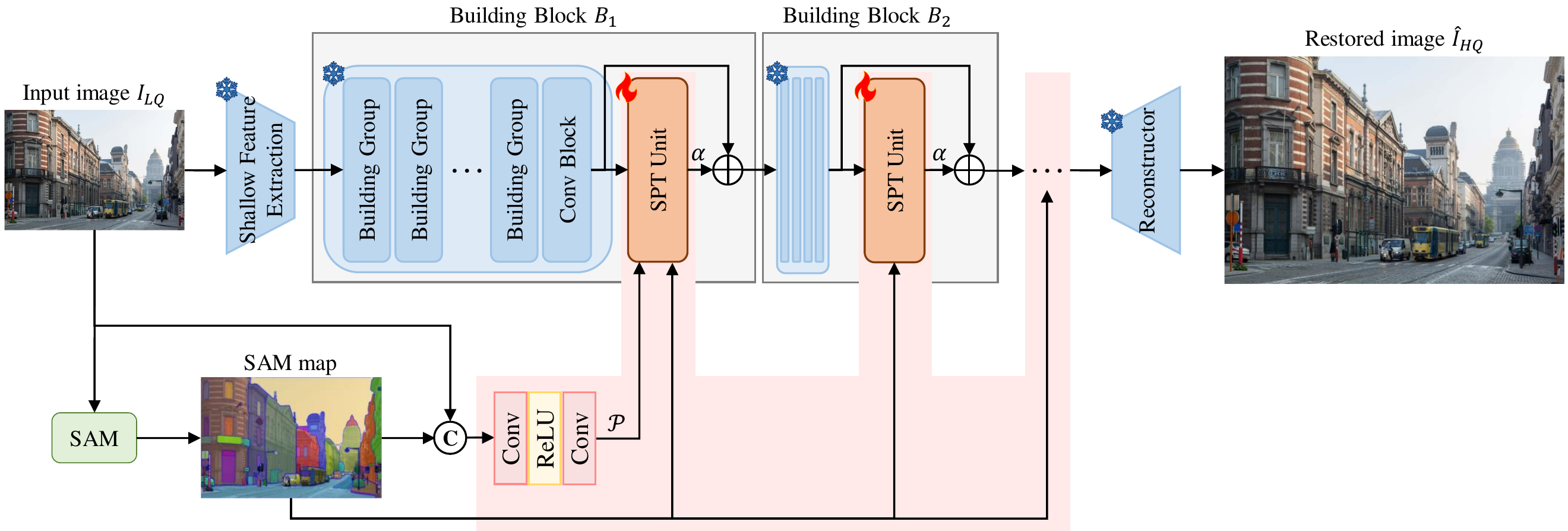}
\caption{Illustration of our proposed method.
In comparison to traditional IR methods that typically employ a shallow feature extractor followed by a deep feature extractor with multiple building blocks and a reconstructor, we present a novel method that efficiently improves network performance by leveraging prior knowledge obtained from SAM~\cite{kirillov2023segment}.
Our proposed method involves integrating semantic masks obtained from SAM into SPT units, which combine the semantic priors with intermediate features of existing IR methods.
As the SPT unit is the only trainable module, our approach is both efficient and scalable compared to full fine-tuning scheme.
Incorporating the SAM prior into our SPT unit allows for effective exploitation of prior knowledge from the large-scale foundation model and improved restoration quality.
}
\label{fig:pipeline}
\end{figure*}

\subsection{Learned Image Priors}
Convolutional neural networks (CNNs)~\cite{dong2015image,dai2019second,mei2020image} have been proposed to capture useful priors by learning mappings between LQ and HQ images from external training data. 
Recent research has shown that deep CNNs can implicitly capture image statistics, making them effective priors for restoring corrupted images.
DIP~\cite{ulyanov2018deep} and single image generative adversarial networks (SinGAN)~\cite{shaham2019singan} have demonstrated the effectiveness of these deep priors in IR tasks, but their applicability may be limited due to their reliance on image-specific statistics.
While other deep priors such as deep denoiser prior~\cite{arjomand2017deep,zhang2017learning}, TNRD~\cite{chen2016trainable}, and LCM~\cite{athar2018latent} have been developed for IR tasks, our focus is not on competing with them.
Instead, we aim to study and exploit the integration of knowledge from large-scale foundation models (\textit{e.g.}, SAM~\cite{kirillov2023segment}) for IR.
To the best of our knowledge, this is the first attempt to leverage the prior knowledge from SAM for IR tasks.
By introducing the prior generated from SAM in a tuning manner, we aim to further improve the performance of existing IR methods without proposing any new architecture.
Our approach complements existing deep priors and provides a promising direction for future research in the field of IR.

\subsection{Large-Scale Foundation Models}
In the era of big data, large-scale foundation models become important components of artificial intelligence.
The recent development of large models mainly benefits from the advanced training schemes (\textit{e.g.}, self-supervised training \cite{hadsell2006dimensionality,doersch2015unsupervised,kenton2019bert}) and scalable network architectures (\textit{e.g.}, Transformer \cite{vaswani2017transformer,dosovitskiy2020ViT}).
The early works such as BERT \cite{kenton2019bert} and RoBERTa \cite{liu2019roberta} utilize masked language modeling to obtain powerful pre-trained models on various natural language processing (NLP) tasks.
Most recently, ChatGPT and GPT-4 \cite{OpenAI2023GPT4TR} developed by OpenAI demonstrates remarkable capabilities on a variety of domains, and even shows sparks of artificial general intelligence \cite{bubeck2023sparks}.
In computer vision, to leverage large-scale image data in a self-supervised manner, contrastive learning \cite{chen2020simple,chen2020improved} and masked image modeling \cite{he2022masked,xie2022simmim} have been explored, which provide rich pre-trained knowledge for downstream tasks.
As a representative work, CLIP \cite{radford2021learning} learns visual representations from the supervision of natural language using 400 million image-text pairs, showing an impressive transferable ability.
Besides, recent works such as IPT \cite{chen2021IPT} and DegAE \cite{liu2023degae} demonstrate foundation models pre-trained on the large-scale data can improve the performance of low-level vision tasks.
Recently, Meta AI Research released a foundation model namely SAM \cite{kirillov2023segment} for open-world image segmentation.
Due to its great potential, an important future direction is to use SAM to aid the downstream tasks \cite{lu2023can}.
In this paper, we explore how to improve IR performance with the semantic prior knowledge from SAM.

\subsection{Parameter-Efficient Fine-tuning}

To introduce additional knowledge from a new dataset or domain into the well-trained models, early works usually fine-tune the whole model parameters \cite{girshick2015fast,he2017mask,he2019rethinking}.
However, this scheme requires a large amount of computational resources and time.
As an alternative, parameter-efficient fine-tuning \cite{lester2021power,li2021prefix,liu2021pre} is firstly proposed in NLP to exploit pre-trained large language model.
It has also been extensively studied for image classification tasks.
For instance, SpotTune \cite{guo2019spottune} studies different fine-tuned layers, TinyTL \cite{cai2020tinytl} only learns the bias modules, and side-tuning \cite{zhang2020side} trains a lightweight network and uses summation to fuse it with the pre-trained network.
Regarding vision and language models, \textit{e.g.}, CLIP \cite{radford2021learning}, parameter-efficient tuning \cite{zhou2022learning,yu2022task} is also leveraged for the performance enhancement on downstream tasks.
Some recent methods such as Adapter \cite{houlsby2019parameter} and VPT \cite{jia2022visual} are developed for Transformer-based architectures, which insert a small number of learnable parameters inside each Transformer layer.
Different from these works, we study the parameter-efficient fine-tuning for IR with the purpose of introducing the semantic prior knowledge.

\section{Preliminary}

\subsection{Network Definition}

For an LQ input image ${I_{LQ}}\in\mathbb{R}^{H \times W \times C_{in}}$, an IR network $\text{IRNet}(\cdot)$ can generate an HQ image $\hat{I}_{HQ}\in\mathbb{R}^{rH \times rW \times C_{out}}$
\begin{equation}
    \hat{I}_{HQ} = \text{IRNet}(I_{LQ}),
\end{equation}
which should be close to the ground-truth image $I_{GT}$.
$H$, $W$, $C_{in}$, $C_{out}$, and $r$ are the image height, width, input channel, output channel, and the scale factor for image super-resolution, respectively.

As shown in Figure~\ref{fig:pipeline}, an IR network consists of three main components: shallow feature extraction, deep feature extraction, and the reconstruction part. 
Without loss of generality, we leverage a convolution layer as shallow feature extraction to get the low-level feature $F_0\in\mathbb{R}^{H \times W \times C}$
\begin{equation}
   F_0 = \text{Enc}(I_{LQ}),
\end{equation}
where $C$ denotes the feature number, and $\text{Enc}(\cdot)$ denotes the convolution layer, serving for the shallow feature extraction.
Then, the shallow feature is processed by the deep feature extraction module, which is composed of $N_1$ building blocks, obtaining the extracted deep feature $F_{DF}\in\mathbb{R}^{H \times W \times C}$.
The above procedure can be formulated as
\begin{equation}
   F_{DF} = {B}_{N_1}(\dots ({B}_2({B}_1 (F_0)) \dots),
\end{equation}
where ${B}_i(\cdot)$ is the $i$-th building block.
Finally, we can get the HQ output image through the reconstruction part
\begin{equation}
   \hat{I}_{HQ} = \text{Rec}(F_{DF}),
\end{equation}
where $\text{Rec}(\cdot)$ denotes the reconstruction part.
In terms of the composition of the reconstruction module, it varies depending on the specific IR task. In the case of image super-resolution, a sub-pixel convolution layer~\cite{shi2016real} with a factor of $r$ is used to upsample the deep feature $F_{DF}$ to match the size of the high-resolution output.
This is followed by a convolution layer both before and after the upsampling module to aggregate the features.
On the other hand, for the tasks such as image denoising, the reconstruction module only consists of a single convolution layer that adjusts the channel dimension of $F_{DF}$ from $C$ to $C_{out}$. 
The LQ input is then added to the convolution output to produce the final output.
This residual learning approach can help accelerate the convergence of the network during training.

\begin{figure*}[!t]
\centering
\includegraphics[width=1\linewidth]{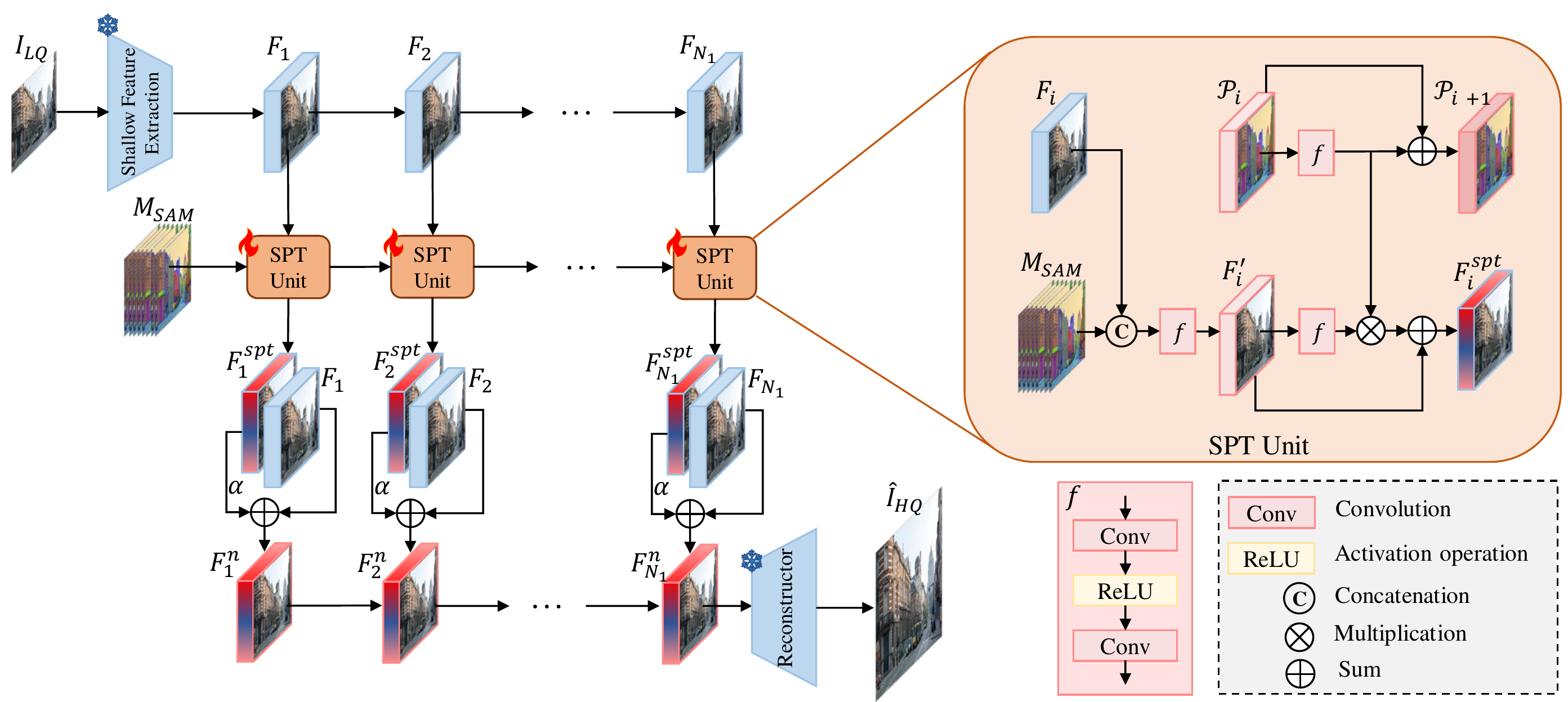}
\caption{Illustration of the SPT unit and the efficient tuning scheme.
The SPT unit takes in the semantic map $M_{SAM}$ extracted from SAM, the deep feature $F_i$ extracted from the $i$-th building block, and the SAM prior representation $\mathcal{P}$ as input.
It then outputs a new feature map $F_i^{spt}$, which incorporates the correlation between $F_i$ and $\mathcal{P}$. 
To efficiently incorporate this new feature map, it is added to the original feature map $F_i$ with a weighting factor of $\alpha$.
The tuned feature maps are then fed into the subsequent building blocks of the network.
}
\label{fig:SPT}
\end{figure*}

\subsection{Segment Anything Model}
In recent years, there has been a growing interest in foundational models pre-trained on large-scale datasets due to their ability to generalize to various downstream tasks.
One such example is the recently released SAM by Meta AI Research~\cite{kirillov2023segment}.
By incorporating a single user prompt, SAM can accurately segment any object in any image or video without the need for additional training, which is commonly referred to as the zero-shot transfer in the computer vision community. According to \cite{kirillov2023segment}, SAM's impressive capabilities are derived from a vision foundation model that has been trained on an extensive SA-1B dataset comprising over 11 million images and one billion masks.

The emergence of SAM has undoubtedly demonstrated strong generalization across various images and objects, opening up new possibilities and avenues for intelligent image analysis and understanding.
Given an image $I\in\mathbb{R}^{H \times W \times C_{in}}$, SAM can generate a segmentation mask tensor $M_{SAM} \in\mathbb{R}^{H \times W \times {N_c}}$
\begin{equation}
   M_{SAM} = \text{SAM}(I),
\end{equation}
where ${N_c}$ denotes the number of masks.
SAM has shown robustness in segmenting low-quality images and producing relatively accurate semantic masks. 
Therefore, we propose to leverage these semantic masks as priors for IR.
By utilizing the rich semantic information in the maps, the IR networks are able to restore more HQ details in the reconstructed images.

We prompt SAM with an
$8\times8$ regular grid of foreground points for each degraded image, resulting in less than 64 masks in most cases.
We fix the number of masks fed into the image restoration networks as 64 by adopting the zero-padding when the masks are insufficient and truncation when the number of masks is larger than 64.
We also discuss more choices of the number of masks in our experimental part.

\section{Method}
\subsection{SAM Prior Tuning Unit}
SAM~\cite{kirillov2023segment} has shown to have promising segmentation capabilities in various scenarios and is robust to various image degradations.
We utilize the extracted semantic map $M_{SAM}$ from SAM as the prior to provide diverse and rich information to improve the performance of existing IR methods.

We first concatenate the LQ input image $I_{LQ}$ and the semantic map $M_{SAM}$ extracted from SAM along the channel dimension.
Then, the concatenated feature is fed to two convolution layers with a ReLU activation operation between them (denoted as $f(\cdot)$), resulting in the SAM prior representation $\mathcal{P} \in \mathbb{R}^{H \times W \times C}$
\begin{equation}
   \mathcal{P} = f([I_{LQ}, M_{SAM}]).
\end{equation}

To provide a concrete example of how the SPT unit works, we use the feature $F_i$ extracted from the $i$-th building block without loss of generality.
As shown in Figure~\ref{fig:SPT}, we first concatenate $F_i^{'}$ with $M_{SAM}$ and feed the concatenated feature to $f(\cdot)$ to generate the enhanced feature representation $F_i^{'}$.
Next, $F_i^{'}$ and $\mathcal{P}$ are separately fed into the feature branch and the SAM prior branch, respectively.
Each branch consists of two convolution layers with ReLU activation in between.
The output features of both branches are multiplied to obtain the correlation, and skip connections are added to both branches to enhance the representation ability of the entire SPT unit.
These procedures can be formulated as
\begin{equation}
\begin{aligned}
   \mathcal{P}_{i+1} = f(\mathcal{P}_i) + \mathcal{P}_i, \\
   F_i^{spt} = f(F_i^{'})*f(\mathcal{P}) + F_i^{'}.
\end{aligned}
\end{equation}
By inserting the SPT unit into $N_1$ building blocks of existing IR networks as a plug-and-play unit, a new network structure is formed, which can utilize the semantic information from the SAM prior to improve IR performance.

\subsection{Efficient Tuning Scheme}
In order to reduce the computational cost during the training stage, we introduce a parameter-efficient tuning scheme that leverages pre-trained IR networks.
Instead of training an IR network from scratch or re-training an existing one, we only update the trainable parameters.
This not only reduces the computational cost but also enhances the overall efficiency of our method.

To incorporate the new feature map $F_i^{spt}$ processed by the SPT unit, we add it to the original feature map $F_i$ using a weighting factor of $\alpha$
\begin{equation}
\begin{aligned}
   F_i^n = F_i + \alpha F_i^{spt}, \\
         = F_i + \alpha \phi_\Theta(F_i),
\end{aligned}
\end{equation}
where $\phi_\Theta(\cdot)$ is the SPT unit $\phi$ with tunable parameters $\Theta$ to the pre-trained IR networks to transform the pre-trained features to new ones.
The incorporation of the enhanced feature map into the original feature map is a straightforward yet powerful operation that allows subsequent building blocks to exploit the semantic information from the SAM prior. By replacing the original feature map with the enhanced one, our proposed approach achieves improved restoration quality without significant computational costs.

In contrast to retraining an entirely new network, our method builds upon existing pre-trained IR networks and only updates the parameters of the SPT units. This parameter-efficient approach significantly reduces the computational burden and makes it a cost-effective solution for improving the performance of existing IR networks.

\begin{table*}[!t]
  \center
  \begin{center}
  \caption{
  Quantitative comparison of baseline methods and their \colorbox{aaa}{SPT unit-tuned variants} in terms of PSNR (dB, $\uparrow$) for the image SR task.
  }
  \label{tab:sr}
  \resizebox{\textwidth}{!}{%
  \begin{tabular}{cccccccccccccc}
  \toprule
  \multirow{2}{*}{Method} 
  & \multirow{2}{*}{Scale} 
  & \multicolumn{2}{c}{Set5} 
  & \multicolumn{2}{c}{Set14} 
  & \multicolumn{2}{c}{B100} 
  & \multicolumn{2}{c}{Urban100} 
  & \multicolumn{2}{c}{Manga109}
  & \multicolumn{2}{c}{Average}
  \\
  \cmidrule{3-14}
  & & PSNR & $\Delta$ 
  & PSNR & $\Delta$
  & PSNR & $\Delta$
  & PSNR & $\Delta$
  & PSNR & $\Delta$
  & PSNR & $\Delta$
  \\
  \midrule
ART  & $\times 2$ & 38.5631  & -                              & 34.5924 & -        & 32.5768  & -       & 34.3001  & -        & 40.2425  & -        & 35.8269  & -        \\
\rowcolor{aaa}ART  & $\times 2$ & 38.5741  & \textbf{\textcolor{citecolor}{+0.0109}}                         & 34.6315 & \textbf{\textcolor{citecolor}{+0.0391}}   & 32.5983  & \textbf{\textcolor{citecolor}{+0.0215}}  & 34.3712  & \textbf{\textcolor{citecolor}{+0.0710}}   & 40.2888  & \textbf{\textcolor{citecolor}{+0.0463}}   & 35.8724  & \textbf{\textcolor{citecolor}{+0.0454}}   \\
ART  & $\times 3$ & 35.0736  & -                              & 31.0183 & -        & 29.5056  & -       & 30.1037  & -        & 35.3889  & -        & 31.7925  & -        \\
\rowcolor{aaa}ART  & $\times 3$ & 35.0919 &  \textbf{\textcolor{citecolor}{+0.0182}}                         & 31.0598 & \textbf{\textcolor{citecolor}{+0.0415}}   & 29.5362  & \textbf{\textcolor{citecolor}{+0.0305}}  & 30.2219  & \textbf{\textcolor{citecolor}{+0.1182}}   & 35.4513  & \textbf{\textcolor{citecolor}{+0.0624}}   & 31.8607  & \textbf{\textcolor{citecolor}{+0.0682}}   \\
ART  & $\times 4$ & 33.0448  & -                              & 29.1585 & -        & 27.9668  & -       & 27.7747  & -        & 32.3081  & -        & 29.4792  & -        \\
\rowcolor{aaa}ART  & $\times 4$ & 33.1113  & \textbf{\textcolor{citecolor}{+0.0665}}                         & 29.2475 & \textbf{\textcolor{citecolor}{+0.0890}}  & 28.0154  & \textbf{\textcolor{citecolor}{+0.0486}}  & 28.1717  & \textbf{\textcolor{citecolor}{+0.3970}}   & 32.5648  & \textbf{\textcolor{citecolor}{+0.2568}}   & 29.7052  & \textbf{\textcolor{citecolor}{+0.2260}}   \\
\midrule
CAT  & $\times 2$ & 38.5079  & -                              & 34.7776 & -        & 32.5853  & -       & 34.2577  & -        & 40.1030  & -        & 35.7773  & -        \\
\rowcolor{aaa}CAT  & $\times 2$ & 38.5230  & \textbf{\textcolor{citecolor}{+0.0151}}                         & 34.8017 & \textbf{\textcolor{citecolor}{+0.0241}}   & 32.5954  & \textbf{\textcolor{citecolor}{+0.0101}}  & 34.2786  & \textbf{\textcolor{citecolor}{+0.0209}}   & 40.1584  & \textbf{\textcolor{citecolor}{+0.0554}}   & 35.8064  & \textbf{\textcolor{citecolor}{+0.0291}}   \\
CAT  & $\times 3$ & 35.0550  & -                              & 31.0433 & -        & 29.5194  & -       & 30.1184  & -        & 35.3838  & -        & 31.8003  & -        \\
\rowcolor{aaa}CAT  & $\times 3$ & 35.0730  & \textbf{\textcolor{citecolor}{+0.0180}}                         & 31.0629 & \textbf{\textcolor{citecolor}{+0.0196}}  & 29.5286  & \textbf{\textcolor{citecolor}{+0.0092}}  & 30.1441  & \textbf{\textcolor{citecolor}{+0.0256}}   & 35.4002  & \textbf{\textcolor{citecolor}{+0.0163}}   & 31.8175  & \textbf{\textcolor{citecolor}{+0.0172}}   \\
CAT  & $\times 4$ & 33.0769  & -                              & 29.1779 & -        & 27.9871  & -       & 27.8861  & -        & 32.3891  & -        & 29.5476  & -        \\
\rowcolor{aaa}CAT  & $\times 4$ & 33.1106 & \textbf{\textcolor{citecolor}{+0.0337}}                       & 29.1995 & \textbf{\textcolor{citecolor}{+0.0216}} & 28.0093 & \textbf{\textcolor{citecolor}{+0.0223}} & 27.8930 & \textbf{\textcolor{citecolor}{+0.0069}} & 32.4817 & \textbf{\textcolor{citecolor}{+0.0926}} & 29.5887 & \textbf{\textcolor{citecolor}{+0.0411}}\\
\midrule
  IMDN & $\times 2$ & 37.9105  & -                              & 33.5949 & -        & 32.1535  & -       & 32.1351  & -        & 38.7899  & -        & 34.5026  & -        \\
\rowcolor{aaa}IMDN & $\times 2$ & 37.8891  & {-0.0215} & 33.6793 & \textbf{\textcolor{citecolor}{+0.0844}}   & 32.1711  & \textbf{\textcolor{citecolor}{+0.0176}}  & 32.2199  & \textbf{\textcolor{citecolor}{+0.0848}}   & 38.9840  & \textbf{\textcolor{citecolor}{+0.1940}}   & 34.6015  & \textbf{\textcolor{citecolor}{+0.0989}}   \\
IMDN & $\times 3$ & 34.3233  & -                              & 30.3066 & -        & 29.0732  & -       & 28.1488  & -        & 33.5833  & -        & 30.4228  & -        \\
\rowcolor{aaa}IMDN & $\times 3$ & 34.3869  & \textbf{\textcolor{citecolor}{+0.0636}}                         & 30.3067 & \textbf{\textcolor{citecolor}{+0.0001}}  & 29.1087  & \textbf{\textcolor{citecolor}{+0.0355}}  & 28.3076  & \textbf{\textcolor{citecolor}{+0.1588}}   & 33.8483  & \textbf{\textcolor{citecolor}{+0.2651}}   & 30.5711  & \textbf{\textcolor{citecolor}{+0.1483}}   \\
IMDN & $\times 4$ & 32.1867  & -                              & 28.5724 & -        & 27.5439  & -       & 26.0318  & -        & 30.4370  & -        & 28.1590  & -        \\
\rowcolor{aaa}IMDN & $\times 4$ & 32.2018  & \textbf{\textcolor{citecolor}{+0.0151}}                         & 28.6088 & \textbf{\textcolor{citecolor}{+0.0364}}   & 27.5814  & \textbf{\textcolor{citecolor}{+0.0374}}  & 26.2896  & \textbf{\textcolor{citecolor}{+0.2578}}   & 30.7284  & \textbf{\textcolor{citecolor}{+0.2913}}   & 28.3476  & \textbf{\textcolor{citecolor}{+0.1886}}   \\
\bottomrule
  \end{tabular}
  } 
  \end{center}
\end{table*}

\begin{table*}[t]
  \center
  \begin{center}
  \caption{
  Quantitative comparison of baseline methods and their \colorbox{aaa}{SPT unit-tuned variants} in terms of PSNR (dB, $\uparrow$) for the color image denoising task.
  }
  \label{tab:dn}
  \resizebox{0.865\textwidth}{!}{%
  \begin{tabular}{cccccccccccc}
  \toprule
  \multirow{2}{*}{Method} 
  & \multirow{2}{*}{$\sigma$ value} 
  & \multicolumn{2}{c}{BSD68} 
  & \multicolumn{2}{c}{Kodak24} 
  & \multicolumn{2}{c}{McMaster} 
  & \multicolumn{2}{c}{Urban100}
  & \multicolumn{2}{c}{Average}
  \\
  \cmidrule{3-12}
  & & PSNR & $\Delta$ 
  & PSNR & $\Delta$
  & PSNR & $\Delta$
  & PSNR & $\Delta$
  & PSNR & $\Delta$
  \\
  \midrule
ART & $15$ &  34.4599  & -      & 35.3871 & -      & 35.6765  & -      & 35.2938  & -      & 35.0672 & -      \\
\rowcolor{aaa}ART & $15$ & 34.4615 & \textbf{\textcolor{citecolor}{+0.0016}} & 35.3921 & \textbf{\textcolor{citecolor}{+0.0050}} & 35.6813 & \textbf{\textcolor{citecolor}{+0.0049}} & 35.2999 & \textbf{\textcolor{citecolor}{+0.0062}} & 35.0717 & \textbf{\textcolor{citecolor}{+0.0044}} \\
\midrule
ART & $25$ & 31.8372  & -      & 32.9526 & -      & 33.4057  & -      & 33.1415  & -      & 32.7202 & -      \\
\rowcolor{aaa}ART& $25$ & 31.9233 & \textbf{\textcolor{citecolor}{+0.0862}} & 33.0058 & \textbf{\textcolor{citecolor}{+0.0532}} & 33.4359 & \textbf{\textcolor{citecolor}{+0.0302}} & 33.1994 & \textbf{\textcolor{citecolor}{+0.0579}} & 32.7844 & \textbf{\textcolor{citecolor}{+0.0642}} \\
\midrule
ART & $50$ & 28.6349  & -      & 29.8659 & -      & 30.3100  & -      & 30.1926  & -      & 29.6609 & -      \\
\rowcolor{aaa}ART & $50$ & 28.6369 & \textbf{\textcolor{citecolor}{+0.0020}} & 29.8674 & \textbf{\textcolor{citecolor}{+0.0015}} & 30.3127 & \textbf{\textcolor{citecolor}{+0.0027}} & 30.2001 & \textbf{\textcolor{citecolor}{+0.0075}} & 29.6656 & \textbf{\textcolor{citecolor}{+0.0046}}\\
  \bottomrule
  \end{tabular}
  } 
  \end{center}
\end{table*}

\section{Experiments}

\subsection{Experimental Settings}
\label{subsec:settings}

\noindent \textbf{Data and Evaluation.} 
We conduct experiments on two typical IR tasks: image SR and color image denoising.
For image SR, we use DIV2K~\cite{timofte2017ntire} and Flickr2K~\cite{lim2017enhanced} as training data, while Set5~\cite{bevilacqua2012low}, Set14~\cite{zeyde2012single}, B100~\cite{martin2001database}, Urban100~\cite{huang2015single}, and Manga109~\cite{matsui2017sketch} are used as test data.
As for color image denoising, we follow the same training data as ART~\cite{zhang2022accurate}, which includes DIV2K, Flickr2K, BSD500~\cite{arbelaez2010contour}, and WED~\cite{ma2016waterloowed}.
We evaluate our proposed method using BSD68~\cite{martin2001database}, Kodak24~\cite{franzen1999kodak}, McMaster~\cite{zhang2011color}, and Urban100 as test data for color image denoising.
The performance is evaluated in terms of PSNR and SSIM~\cite{wang2004image} values on the Y channel of images transformed to the YCbCr space for image SR, and on the RGB channel for color image denoising.

\noindent \textbf{Selected baseline methods.} 
To evaluate the effectiveness of our proposed method, we conduct experiments using several representative methods in two IR tasks.
For image SR, we select three representative methods: IMDN~\cite{hui2019lightweight}, a typical light-weight CNN-based SR method, as well as the state-of-the-art vision Transformer-based methods ART~\cite{zhang2022accurate} and CAT~\cite{chen2022cross}. 
 
\noindent \textbf{Training Settings.} 
Data augmentation is performed on the training data through horizontal flip and random rotation of $90^{\circ}$, $180^{\circ}$, and $270^{\circ}$.
Besides, we crop the original images into 64$\times$64 patches as the basic training inputs for image SR and 128$\times$128 patches for image denoising.
We add the SPT units after each buliding block, and the batch size is set to 4.
We choose ADAM~\cite{kingma2014adam} to optimize the networks with $\beta_1=0.9$, $\beta_2=0.999$, and zero weight decay.
The initial learning rate is set as 1$\times 10^{-4}$.
We fine-tune the parameters of ART, CAT, and IMDN until convergence, and we adjust the learning rate to half every 5,000 iterations.
Experiments are conducted with a single NVIDIA 3090 GPU.

\begin{figure*}[!t]
\centering
\includegraphics[width=\linewidth]{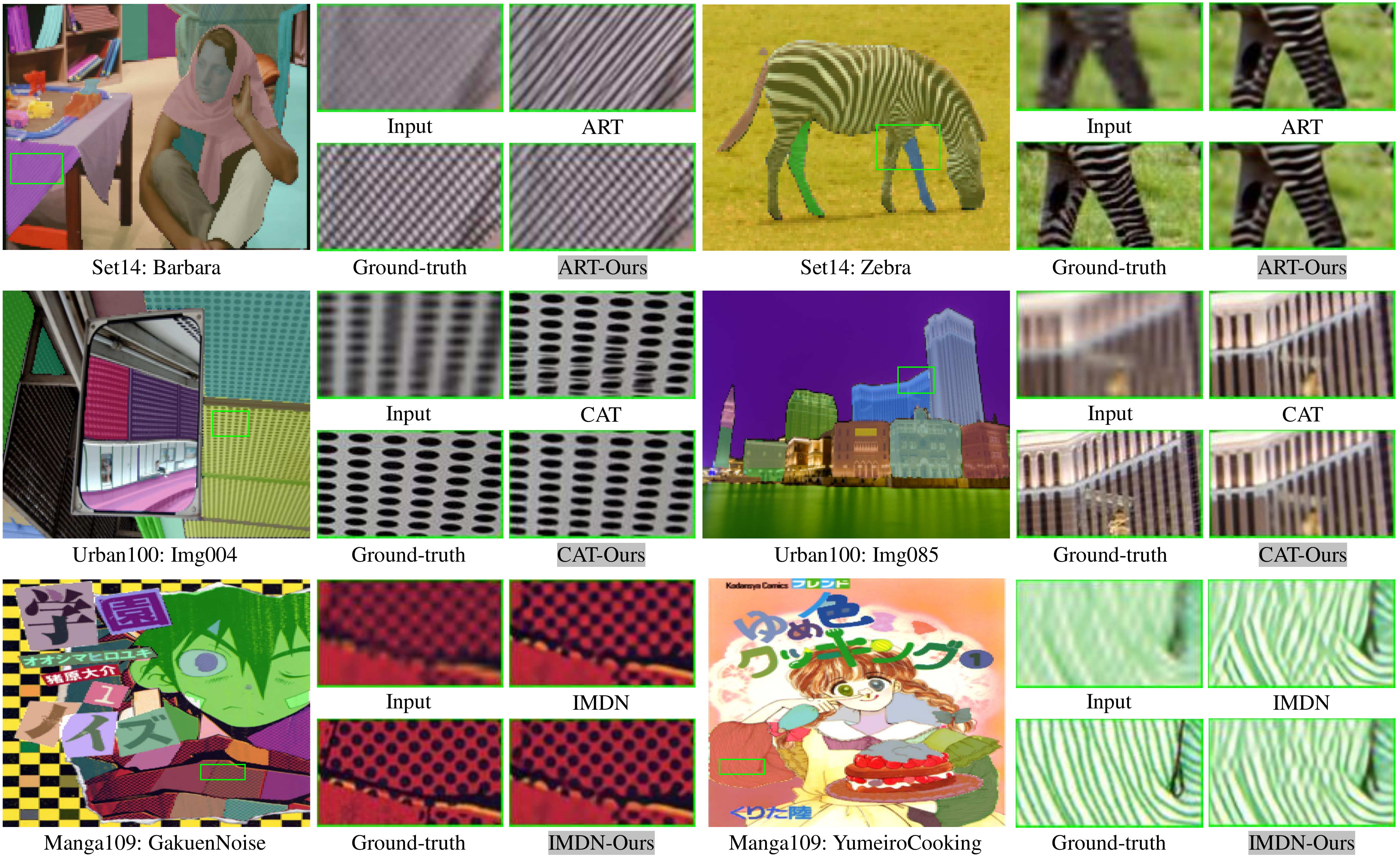}
\caption{Visual comparisons on $\times 4$ image super-resolution.
We show the results of extracted SAM masks, input LQ images, the ground-truth HQ images, the baseline methods, and the baseline methods trained with our proposed method.
}
\label{fig:sr}
\end{figure*}

\begin{figure*}[!t]
\centering
\includegraphics[width=\linewidth]{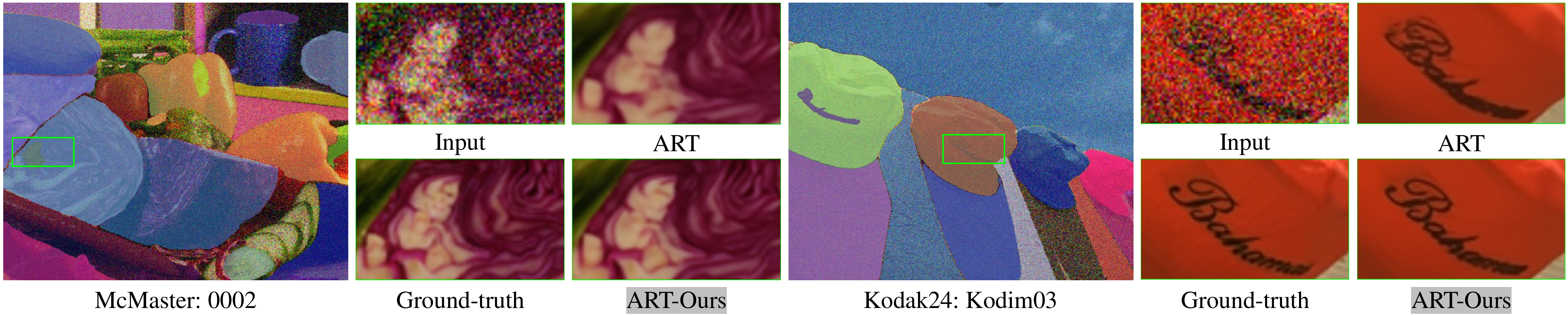}
\caption{Visual comparisons on color image denoising ($\sigma=50$).
We show the results of extracted SAM masks, input LQ images, the ground-truth HQ images, ART, and ART trained with our proposed method.
}
\label{fig:dn}
\end{figure*}

\subsection{Quantitative and Qualitative Comparisons}
We evaluate the effectiveness of our proposed method by comparing representative baseline methods and their SPT unit tuned versions on the tasks of image SR and color image denoising.

\noindent \textbf{Image super-resolution.}
Table~\ref{tab:sr} presents a quantitative comparison between methods trained with and without the SPT unit on benchmark datasets for image SR.
The results show that the existing image super-resolution methods fine-tuned with SPT units outperform the corresponding baselines by a significant margin.
For example, in the $\times 4$ super-resolution of Urban100 dataset, ART fine-tuned with our proposed method achieves 28.1717dB (PSNR), while the same baseline network only achieves 27.7747dB (PSNR).
The weighted average values in the table demonstrate that our method effectively utilizes the SAM prior, leading to further performance improvements in existing SR methods.

Figure~\ref{fig:sr} illustrates visual comparisons of SR results obtained by the baseline methods and their tuned versions. 
We observe that the existing SR methods tend to generate realistic detailed textures but with visual aliasing and artifacts.
For example, in the first example of Figure~\ref{fig:sr}, ART produces blurry details of the tablecloths.
On the other hand, ART tuned with our proposed method reconstructs sharp and natural details.
This indicates that our method effectively employs semantic priors to capture the characteristics of each category, leading to more natural and realistic textures.
This observation is consistent with the approach presented in \cite{wang2018sft}.

\noindent \textbf{Color image denoising.}
Table~\ref{tab:dn} presents quantitative comparisons for color image denoising.
The results show that ART fine-tuned with SPT units outperforms the original ART by a significant margin on three different levels of noise.
For instance, in the $\sigma=25$ color image denoising task, ART fine-tuned with our proposed method achieves an average PSNR of 32.7844dB, which is 0.0642dB higher than the same baseline network.
As shown in Figure~\ref{fig:dn}, the color image denoising results of ART fine-tuned with our method exhibit better visual quality than the original ART.
The images restored by our method have more details and fewer blocking artifacts, leading to sharper edges and more explicit textures.
These results demonstrate that our method can effectively leverage semantic priors to improve the performance of existing color image denoising methods.

\subsection{Ablation Study}
\label{subsec:ablation}
For the ablation study, we use the dataset DIV2K~\cite{timofte2017ntire} and Flickr2K~\cite{lim2017enhanced} to train ART on the $\times 4$ image SR task.
The results are evaluated on the dataset of Manga109.

\noindent \textbf{The effectiveness of the SPT Unit}.
To evaluate the effectiveness of the proposed SPT unit, we design several variants as follows:
(1) SPT-$F_i$: we feed $M_{SAM}$ directly to $f(\cdot)$ without concatenating it with $F_i$;
(2) SPT-$P_i$: we remove the extracted SAM prior representation $\mathcal{P}_i$ from the SPT unit;
(3) SPT-cat: we concatenate $M_{SAM}$, $F_i$, and $\mathcal{P}_i$ and feed the concatenated tensor to $f(\cdot)$, generating $F_i^{spt}$.
The corresponding results are shown in Table~\ref{tab:sptablation}, where it can be observed that although these variants achieve some performance improvements, they are far less effective than our designed SPT unit.
This indicates that our SPT unit is simple yet effective, and can better utilize the semantic prior information from the SAM mask for image SR.
We also analyze the effect of inserting the SPT unit at different positions on the final performance.
Table~\ref{tab:sptablation} shows the results.
It can be observed that as the number of SPT units inserted increases, the final performance gradually improves, and the more units inserted, the more significant the improvement.
For example, when we only insert the SPT unit in the first building block, we only achieve a 0.0141dB improvement.
However, when we insert the SPT unit in all building blocks, we achieve a significant improvement of up to 0.2568dB.

\begin{table}[!t]
  \center
  \begin{center}
  \caption{
  The effectiveness of the SPT unit in different variants and different positions.
  $B_{N_i}$ here denotes the insertion of the SPT units into the building blocks $B_{1}$ to $B_{N_i}$.
  }
  \label{tab:sptablation}
  \resizebox{\linewidth}{!}{%
  \begin{tabular}{cccccc}
  \toprule
\multicolumn{2}{c}{SPT variants} & \multicolumn{4}{c}{SPT locations} \\
  \cmidrule(r){1-2} \cmidrule(r){3-6}
Method     & PSNR            & Block      & PSNR      & Block     & PSNR     \\
\midrule
SPT-$F_i^{'}$     & $32.4694_{+0.1613}$            & $B_1$      & $32.3222_{+0.0141}$      & $B_4$      & $32.4266_{+0.1185}$      \\
SPT-$P_i$            & $32.4519_{+0.1438}$             & $B_2$      & $32.3188_{+0.0107}$      & $B_5$      & $32.4607_{+0.1526}$   \\ 
SPT-cat            & $32.4194_{+0.1113}$             & $B_3$      & $32.4149_{+0.1068}$      & $B_6$      & $32.5648_{+0.2568}$    \\ 
\bottomrule
\end{tabular}
}
  \end{center}
\end{table}

\begin{table}[t]
  \center
  \begin{center}
  \caption{
  Impact of different $\alpha$ values.
  }
  \label{tab:alpha}
  \resizebox{\linewidth}{!}{%
  \begin{tabular}{p{1.5cm}<{\centering}p{2.3cm}<{\centering}p{1.5cm}<{\centering}p{2.3cm}<{\centering}}
  \toprule
$\alpha$    & PSNR            & $\alpha$      & PSNR        \\
\midrule
$\alpha=0.5$    & $32.4332_{+0.1316}$            & $\alpha=1$        & $32.5648_{+0.2568}$    \\
$\alpha=1.5$       & $32.4653_{+0.0995}$            & $\alpha=2$       & $32.4025_{+0.1623}$    \\ 
\bottomrule
\end{tabular}
}
  \end{center}
\end{table}

\begin{table}[t]
  \center
  \begin{center}
  \caption{
  Comparison of different tuning schemes.
  }
  \label{tab:tune}
  \resizebox{0.9\linewidth}{!}{%
  \begin{tabular}{p{2.5cm}<{\centering}p{3cm}<{\centering}p{1.8cm}<{\centering}}
  \toprule
Scheme   & PSNR            & \# Iterations          \\
\midrule
Ours   & $32.5648_{+0.2568}$          & $\sim$8,000    \\
Full fine-tuning     & $32.5640_{+0.2538}$            & $\sim$15,000   \\ 
\bottomrule
\end{tabular}
}
  \end{center}
\end{table}

\begin{table}[t]
  \center
  \begin{center}
  \caption{
  Effectiveness of the extracted SAM masks
  }
  \label{tab:sam}
  \resizebox{0.8\linewidth}{!}{%
  \begin{tabular}{p{4cm}<{\centering}p{3cm}<{\centering}}
  \toprule
SAM mask/representation   & PSNR                \\
\midrule
Coarse   & $32.5648_{+0.2568}$        \\
Medium    & $32.5709_{+0.2628}$    \\ 
Fine    & $32.5737_{+0.2656}$    \\
\bottomrule
\end{tabular}
}
  \end{center}
\end{table}

\noindent \textbf{The effectiveness of the efficient tuning scheme}.
We first conduct an analysis of the impact of different $\alpha$ values on the results.
We select several typical $\alpha$ values (\textit{i.e.}, 0.5, 1.0, 1.5, and 2.0) and compare their effects, as shown in Table~\ref{tab:alpha}.
From the results in Table 1, it can be observed that the best performance is achieved when $\alpha=1.0$.
When $\alpha$ is too large or too small, the weight tuning of the SPT unit cannot be balanced well, leading to sub-optimal performance.
We also compare our tuning method with full-parameter tuning.
As shown in Table~\ref{tab:tune}, our tuning method can improve the performance of the ART network faster and better than the latter.
This is because we base our method on the pre-trained and frozen ART parameters and focused on updating the tuning-related parameters, which enables efficient updates on a small number of parameters.

\noindent \textbf{The effect of the granularity of SAM masks}. 
We adjust the density of the regular grid used to prompt SAM and obtain different groups of masks, 
Usually, a denser grid results in a larger number of masks containing more fine-grained objects. Specifically, we prompt SAM using $8\times8$, $16\times16$, and $24\times24$ grids, which are denoted as $Coarse$, $Medium$, and $Fine$, respectively.
For these three cases, we fix the number of masks fed into the image restoration networks as 64, 128, and 256, respectively, using padding or truncation. In terms of the network architecture, we only adjust the number of the input channel of the first convolutional layer. 
Table~\ref{tab:sam} shows the impact of the granularity of SAM masks on the final results. 
It can be observed that using more masks can improve the performance of ART, which indicates that leveraging more fine-grained semantic information is more helpful and further confirms the effectiveness of the SAM prior.

\begin{figure}[!t]
\centering
\includegraphics[width=0.95\linewidth]{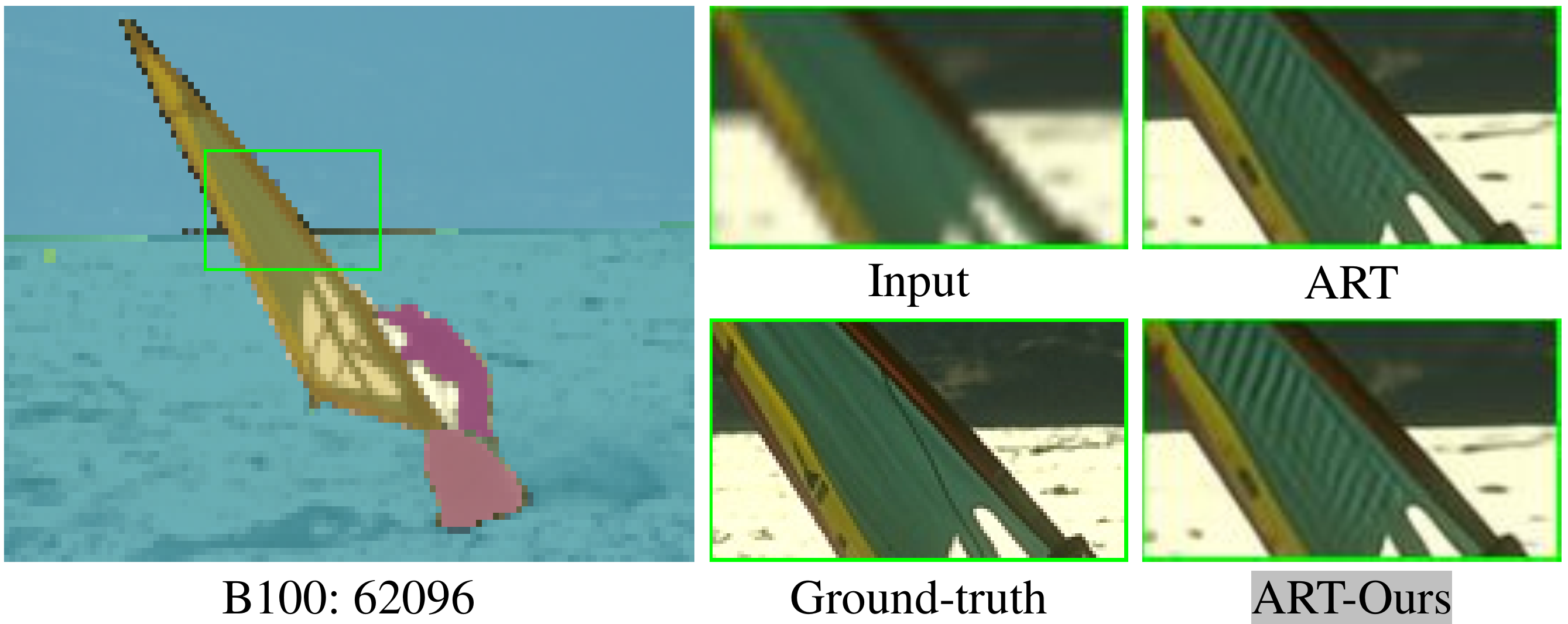}
\caption{
A failure case.
The use of extracted SAM masks as semantic priors in our method can introduce unrealistic fine-grained structures and texture characteristics, resulting in artifacts that deviate significantly from the real one.
}
\label{fig:limit}
\end{figure}

\subsection{Limitations}

This section presents the limitation of our method that arises from the use of extracted SAM masks as semantic priors.
Despite the performance improvement on SR, they may also generate unrealistic fine-grained structures and textures that do not exist in the ground-truth image.
For example, in the sailboat shown in Figure~\ref{fig:limit}, the SAM masks indicate a semantic mask of the sail area, resulting in a grid-like structure that is not present in the ground-truth image.
While this structure may appear visually pleasing to humans, it deviates significantly from the actual image and can be considered as artifacts. To address this limitation, future work could explore more effective methods for incorporating semantic priors into IR tasks.
This could be achieved by investigating different ways to introduce semantic priors into existing methods to improve the fidelity of the generated image.

\section{Conclusion}
In this paper, we propose a novel approach for image restoration that leverages the prior knowledge of the state-of-the-art segment anything model (SAM) to improve the quality of restored images. By incorporating semantic priors extracted from SAM using a light-weight SAM prior tuning (SPT) unit, we significantly enhance the restoration quality. Notably, the SPT unit is the only trainable module in our approach, making it both efficient and scalable. Our experiments demonstrate the effectiveness of the SPT unit as a plug-in to enhance a variety of methods for image super-resolution and denoising. 
More importantly, our work highlights the potential of integrating prior knowledge from large-scale foundation models for improving the performance of image restoration.

\normalem
{\small
\bibliographystyle{ieee}
\bibliography{bib}
}

\end{document}